\documentclass{article}
\usepackage{ijcai21}

\usepackage{times}
\usepackage{soul}
\usepackage{url}
\usepackage[hidelinks]{hyperref}
\usepackage[utf8]{inputenc}
\usepackage[small]{caption}
\usepackage{graphicx}
\usepackage{amsmath}
\usepackage{amsthm}
\usepackage{booktabs}
\usepackage{algorithm}
\usepackage{algorithmic}
\usepackage{subfigure}
\usepackage{multirow}
\usepackage{bm}
\usepackage{color}

\title{HifiFace: 3D Shape and Semantic Prior Guided High Fidelity Face Swapping} 

\author{
Yuhan Wang$^{1,2}$\footnotemark[1] \and
Xu Chen$^{1,3}$\footnotemark[1] \and
Junwei Zhu$^{1}$\and
Wenqing Chu$^1$ \and
Ying Tai$^1$\footnotemark[2] \\
Chengjie Wang$^1$ \and
Jilin Li$^1$ \and
Yongjian Wu$^1$ \and
Feiyue Huang$^1$ \And
Rongrong Ji$^{3,4}$ \\
\affiliations
{ {$^1$Youtu Lab, Tencent}} \\
{ {$^2$Zhejiang University}} \\
{ {$^3$Media Analytics and Computing Lab, Department of Artificial Intelligence, School of Informatics,
Xiamen University}} \\
{ {$^4$Institute of Artiﬁcial Intelligence, Xiamen University}} \\
\emails
{
{~\color{magenta} ~\url{https://johann.wang/HifiFace}}
}
}

\begin{document}

\twocolumn[{%
\renewcommand\twocolumn[1][]{#1}%
\maketitle 

\begin{figure}[H] 
\hsize=\textwidth
\centering
\includegraphics[width=0.95\textwidth]{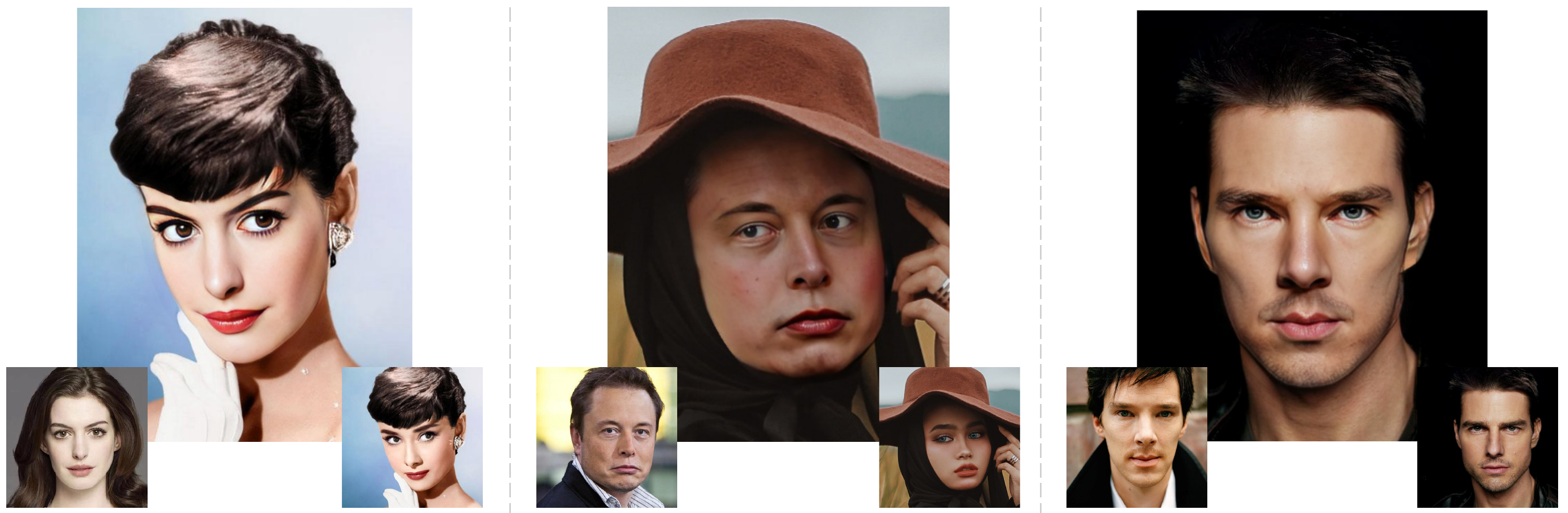}
\caption{Face swapping results generated by our HifiFace. 
The face in the target image is replaced by the face in the source image.} 
\label{fig:title} 
\end{figure}
}]

\footnotetext[1]{Equal Contribution. Work done when Yuhan and Xu are both interns at Youtu Lab, Tencent.} 
\footnotetext[2]{Corresponding Author.}

\begin{abstract}
In this work, we propose a high fidelity face swapping method, called HifiFace, which can well preserve the face shape of the source face and generate photo-realistic results. Unlike other existing face swapping works that only use face recognition model to keep the identity similarity, we propose 3D shape-aware identity to control the face shape with the geometric supervision from $3$DMM and $3$D face reconstruction method. Meanwhile, we introduce the Semantic Facial Fusion module to optimize the combination of encoder and decoder features and make adaptive blending, which makes the results more photo-realistic. Extensive experiments on faces in the wild demonstrate that our method can preserve better identity, especially on the face shape, and can generate more photo-realistic results than previous state-of-the-art methods. 
\end{abstract}

\section{Introduction}

Face swapping is a task of generating images with the identity from a source face and the attributes (\textit{e.g.}, pose, expression, lighting, background, \textit{etc.}) from a target image (as shown in Figure~\ref{fig:title}), which has attracted much interest with great potential usage in film industry~\cite{alexander2009creating} and computer games.

In order to generate high-fidelity face swapping results, there are several critical issues: ($1$) The identity of the result face including the \textit{face shape} should be close to the source face.   
($2$) The results should be \textit{photo-realistic} which are faithful to the expression and posture of the target face and consistent with the details of the target image like lighting, background, and occlusion.

To preserve the identity of the generated face, previous works ~\cite{nirkin2018face,Nirkin2019fsgan,jiang2020deeperforensics} generated inner face region via $3$DMM fitting or face landmark guided reenactment and blend it into the target image, as shown in Figure~\ref{fig:previous_work}(a). These methods are weak in identity similarity because $3$DMM can not imitate the identity details and the target landmarks contain the identity of target image. Also, the blending stage restricts the change of face shape. As shown in Figure~\ref{fig:previous_work}(b),~\cite{liu2019identity,chen2020simswap} draw support from a face recognition network to improve the identity similarity.
However, face recognition network focuses more on texture and is insensitive to the geometric structure. Thus, these methods can not preserve the exact face shape robustly.

As for generating photo-realistic results,~\cite{nirkin2018face,Nirkin2019fsgan} used Poisson blending to fix the lighting, but it tended to cause ghosting and could not deal with complex appearance conditions.~\cite{jiang2020deeperforensics,zhu2020aot,li2019faceshifter} designed an extra learning-based stage to optimize the lighting or occlusion problem, but they are fussy and can not solve all problems in one model. 

To overcome the above defects, we propose a novel and elegant end-to-end learning framework, named HifiFace, to generate high fidelity swapped faces via $3$D shape and semantic prior. 
Specifically, we first regress the coefficients of source and target face by a $3$D face reconstruction model and recombine them as shape information. Then we concatenate it with the identity vector from a face recognition network. We explicitly use the $3$D geometric structure information and use the recombined 3D face model with source's identity, target's expression, and target's posture as auxiliary supervision to enforce precise face shape transfer. With this dedicated design, our framework can achieve more similar identity performance, especially on face shape.

Furthermore, we introduce a Semantic Facial Fusion (SFF) module to make our results more photo-realistic. 
The attributes like lighting and background require spatial information and the high image quality results need detailed texture information. The low-level feature in the encoder contains spatial and texture information, but also contains rich identity from the target image. 
Hence, to better preserve the attributes without the harm of identity, our SFF module integrates the low-level encoder features and the decoder features by the learned adaptive face masks.
Finally, in order to overcome the occlusion problem and achieve perfect background, we blend the output to the target by the learned face mask as well.
Unlike~\cite{Nirkin2019fsgan} that used the face masks of the target image for directly blending, HifiFace learns face masks at the same time under the guidance of dilated face semantic segmentation, which helps the model focus more on the facial area and make adaptive fusion around the edge.  
HifiFace handles image quality, occlusion, and lighting problems in \textit{one} model, making the results more photo-realistic. Extensive experiments demonstrate that our results surpass other State-of-the-Art (SOTA) methods on wild face images with large facial variations. 

Our contributions can be summarized as follows:
\begin{enumerate}
    \item We propose a novel and elegant end-to-end learning framework, named HifiFace, which can well preserve the face shape of the source face and generate high fidelity face swapping results.
    \item We propose a 3D shape-aware identity extractor, which can generate identity vector with exact shape information to help preserve the face shape of the source face.
    \item We propose a semantic facial fusion module, which can solve occlusion and lighting problems and generate results with high image quality.

\end{enumerate}

\section{Related Work}

\begin{figure*}[t] 
\begin{center} 
\includegraphics[width=0.85\linewidth]{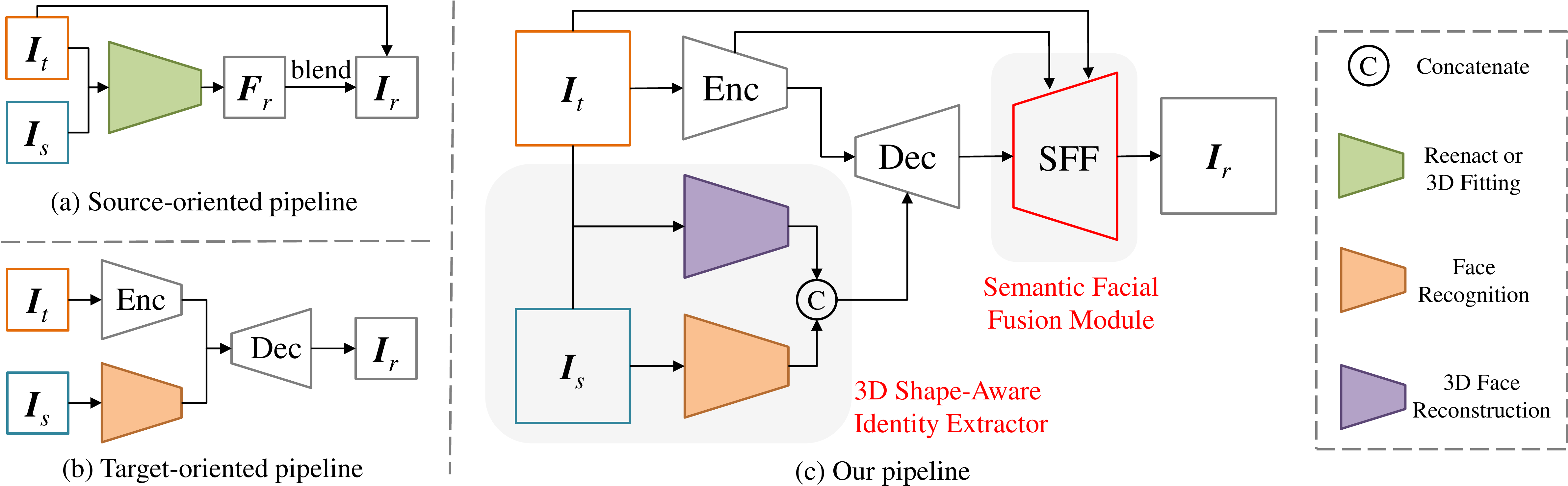} 
\end{center} 
\vspace{-2mm}
\caption{The pipelines of previous works and our HifiFace. 
(a) Source-oriented pipeline uses $3$D fitting or reenactment to generate inner face region and blend it into the target image, in which $\textit{F}_r$ means the face region of the result. 
(b) Target-oriented pipeline uses a face recognition network to exact identity and combines encoder feature with identity in the decoder. 
(c) Our pipeline consists of four parts: the Encoder part, Decoder part, $3$D shape-aware identity extractor, and SFF module. 
The encoder extracts features from $I_t$, and the decoder fuses the encoder feature and the $3$D shape-aware identity feature. Finally, the SFF module helps further improve the image quality.} 
\label{fig:previous_work} 
\end{figure*}

\paragraph{$3$D-based Methods.} 
$3$D Morphable Models ($3$DMM) transformed the shape and texture of the examples into a vector space representation ~\cite{blanz1999morphable}.   
~\cite{thies2016face2face} transferred expressions from source to target face by fitting a 3D morphable face model to both faces. 
 ~\cite{nirkin2018face} transferred the expression and posture by $3$DMM and trained a face segmentation network to preserve the target facial occlusions. 
These $3$D-based methods follow a source-oriented pipeline like Figure~\ref{fig:previous_work}(a) which only generates the face region by $3$D fitting and blends it into the target image by the mask of the target face. They suffer from unrealistic texture and lighting because the $3$DMM and the renderer can not simulate complex lighting conditions. Also, the blending stage limits the face shape. In contrast, our HifiFace accurately preserves the face shape via geometric information of $3$DMM and achieves realistic texture and attributes via semantic prior guided recombination of both encoder and decoder feature.

\paragraph{GAN-based Methods.}
GAN has shown great ability in generating fake images since it was proposed by ~\cite{goodfellow2014generative}.~\cite{isola2017image} proposed a general image-to-image translation method, which proves the potential of conditional GAN architecture in swapping face, although it requires paired data. 

The GAN-based face swapping methods mainly follow source-oriented pipeline or target-oriented pipeline. ~\cite{Nirkin2019fsgan,jiang2020deeperforensics} followed source-oriented pipeline in Figure~\ref{fig:previous_work}(a) which used face landmarks to compose face reenactment. But it may bring weak identity similarity, 
and the blending stage limited the change of face shape. ~\cite{liu2019identity,chen2020simswap,li2019faceshifter} followed target-oriented pipeline in Figure~\ref{fig:previous_work}(b) which used a face recognition network to extract the identity and use decoder to fuse the encoder feature with identity, but they could not robustly preserve exact face shape and is weak in image quality.
Instead, HifiFace in Figure~\ref{fig:previous_work}(c) replaces the face recognition network with a $3$D shape-aware identity extractor to better preserve identity including the face shape and introduces an SFF module after the decoder to further improve the realism.

Among these, FaceShifter~\cite{li2019faceshifter} and SimSwap~\cite{chen2020simswap} follow target-oriented pipeline and can generate high fidelity results. FaceShifter~\cite{li2019faceshifter} leveraged a two-stage framework and achieved state-of-the-art identity performance. But it could not perfectly preserve the lighting despite using an extra fixing stage. However, HifiFace can well preserve lighting and identity in \textit{one} stage. Meanwhile, HifiFace can generate photo-realistic results with higher quality than FaceShifter.  
~\cite{chen2020simswap} proposed weak feature matching loss to better preserve the attributes, but it harms the identity similarity. While HifiFace can better preserve attributes and do not harm the identity.  

\section{Approach}
Let $I_s$ be the source images and $I_t$ the target images, respectively. We aim to generate result image $I_r$ with the identity of the $I_s$ and the attributes of $I_t$. As illustrated in Figure~\ref{fig:previous_work}(c), our pipeline consists of four parts: the Encoder part, Decoder part, $3$D shape-aware identity extractor (Sec.~\ref{sec:sid}), and SFF module (Sec.~\ref{sec:sff}). 
First, we set $I_t$ as the input of the encoder and use several res-blocks~\cite{he2016deep} to get the attribute feature. Then, we use the $3$D shape-aware identity extractor to get $3$D shape-aware identity. 
After that, we use res-block with adaptive instance normalization~\cite{karras2019style} in decoder to fuse the $3$D shape-aware identity and attribute feature. 
Finally, we use the SFF module to get higher resolution and make the results more photo-realistic.

\subsection{3D Shape-Aware Identity Extractor}\label{sec:sid}
Most GAN-based methods only use a face recognition model to obtain identity information in the face swapping task. However, the face recognition network focuses more on texture and is insensitive to the geometric structure.
To get more exact face shape features, we introduce $3$DMM and use a pre-trained state-of-the-art $3$D face reconstruction model~\cite{deng2019accurate} as a shape feature encoder, which represents the face shape $\textbf{S}$ by an affine model:
\begin{align}
   \textbf{S} = \textbf{S}({\alpha},{\beta}) = \bar{\textbf{S}} + \textbf{B}_{id}{\alpha} +  \textbf{B}_{exp}{\beta},
\end{align}%
where $\bar{\textbf{S}}$ is the average face shape; $\textbf{B}_{id}$, $\textbf{B}_{exp}$ are the PCA bases of identity and expression; {$\alpha$} and {$\beta$} are the corresponding coefficient vectors for generating a 3D face.

\begin{figure*}[t] 
\begin{center} 
\includegraphics[width=0.9\linewidth]{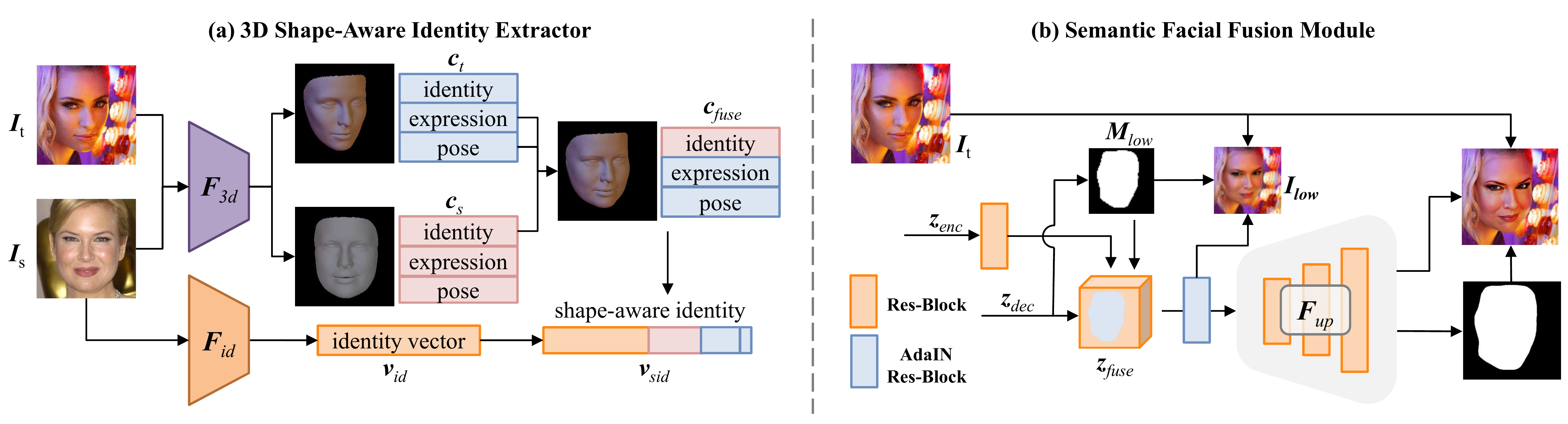} 
\end{center} 
\vspace{-3mm}
\caption{Details of $3$D shape-aware identity extractor and SFF module. 
(a) $3$D shape-aware identity extractor uses $\boldsymbol{F}_{3d}$ ($3$D face reconstruction network) and $\boldsymbol{F}_{id}$ (face recognition network) to generate shape-aware identity. 
(b) SFF module recombines the encoder and decoder feature by $\boldsymbol{M}_{low}$ and makes the final blending by $\boldsymbol{M}_{r}$. The $\boldsymbol{F}_{up}$ means the upsample Module.} 
\label{fig:pipline} 
\end{figure*}

As illustrated in Figure~\ref{fig:pipline}(a), we regress $3$DMM coefficients $\boldsymbol{c}_s$ and $\boldsymbol{c}_t$, containing identity, expression, and posture of the source and target face by the $3$D face reconstruction model $\boldsymbol{F}_{3d}$. 
Then, we generate a new $3$D face model by $\boldsymbol{c}_{fuse}$ with the source's identity, target's expression and posture.
Note that posture coefficients do not decide face shape but may affect the $2$D landmarks locations when computing the loss. 
We do not use the texture and lighting coefficients because the texture reconstruction still remains unsatisfactory. 
Finally, we concatenate the $\boldsymbol{c}_{fuse}$ with the identity feature $\boldsymbol{v}_{id}$ extracted by $\boldsymbol{F}_{id}$, a pre-trained state-of-the-art face recognition model~\cite{huang2020curricularface}, and get the final vector $\boldsymbol{v}_{sid}$, called 3D shape-aware identity. 
Thus, HifiFace achieves well identity information including geometric structure, which helps preserve the face shape of the source image.

\subsection{Semantic Facial Fusion Module}\label{sec:sff}

\paragraph{Feature-Level.} 
The low-level feature contains rich spatial information and texture details, which may significantly help generate more photo-realistic results.
Here, we propose the SFF module to not only make full use of the low-level encoder and decoder features, but also overcome the contradiction in avoiding harming the identity because of the target's identity information in low-level encoder feature.

As shown in Figure~\ref{fig:pipline}(b), we first predict a face mask $\boldsymbol{M}_{low}$ when the decoder features $\boldsymbol{z}_{dec}$ are of size $1/4$ of the target. Then, we blend $\boldsymbol{z}_{dec}$ by $\boldsymbol{M}_{low}$ and get $\boldsymbol{z}_{fuse}$, formulated as:
\begin{align}
   \boldsymbol{z}_{fuse} = \boldsymbol{M}_{low}\odot \boldsymbol{z}_{dec} + (1- \boldsymbol{M}_{low})\odot \sigma(\boldsymbol{z}_{enc}),
\end{align}%
where $\boldsymbol{z}_{enc}$ means the low-level encoder feature with $1/4$ size of the original and $\sigma$ means a res-block~\cite{he2016deep}.

The key design of SFF is to adjust the attention of the encoder and decoder, which helps disentangle identity and attributes.
Specifically, the decoder feature in non-facial area can be damaged by the inserted source's identity information, thus we replace it with the clean low-level encoder feature to avoid potential harm.
While the facial area decoder feature, which contains rich identity information of the source face, should not be disturbed by the target, therefore we preserve the decoder feature in the facial area. 

After the feature-level fusion, we generate $\boldsymbol{I}_{low}$ to compute auxiliary loss for better disentangling the identity and attributes. Then we use a $4\times$ Upsample Module $\boldsymbol{F}_{up}$ which contains several res-blocks to better fuse the feature maps. Based on $\boldsymbol{F}_{up}$, it's convenient for our HifiFace to generate even higher resolution results (\textit{e.g.}, $512\times512$).

\paragraph{Image-Level.}  In order to solve the occlusion problem and better preserve the background, previous works~\cite{Nirkin2019fsgan,natsume2018fsnet} directly used the mask of the target face. 
However, it brings artifacts because the face shape may change.
Instead, we use SFF to learn a sightly dilated mask and embrace the change of the face shape.  
Specifically, we predict a $3$-channel $\boldsymbol{I}_{out}$ and $1$-channel $\boldsymbol{M}_{r}$, and blend $\boldsymbol{I}_{out}$ to the target image by $\boldsymbol{M}_{r}$, formulated as:
\begin{align}
   \boldsymbol{I}_{r} = \boldsymbol{M}_{r}\odot \boldsymbol{I}_{out} + (1- \boldsymbol{M}_{r})\odot \boldsymbol{I}_{t}.
\end{align}%
In summary, HifiFace can generate photo-realistic results with high image quality and well preserve lighting and occlusion with the help of the SFF module. 
Note that these abilities still work despite the change of face shape, because the masks have been dilated and our SFF benefits from inpainting around the contour of predicted face. 

\subsection{Loss Function}

\paragraph{3D Shape-Aware Identity (SID) Loss.} 
SID loss contains shape loss and ID loss. 
We use $2$D landmark keypoints as geometric supervision to constrain the face shape, which is widely used in $3$D face reconstruction~\cite{deng2019accurate}. 
First we use a mesh renderer to generate $3$D face model by coefficients of source image identity and target image expression and posture. 
Then, we generate $3$D face model of $\boldsymbol{I}_r$ and $\boldsymbol{I}_{low}$ by regressing $3$DMM coefficients. 
Finally we project the $3$D facial landmark vertices of reconstructed face shapes
onto the image obtaining landmarks \{{$q^{fuse}$}\}, \{{$q^r$}\} and \{{$q^{low}$}\}:
\begin{align}
   \mathcal{L}_{shape} = & \frac{1}{N}\sum_{n=1}^N ||q^{fuse}_n - q^r_n||_1+||q^{fuse}_n - q^{low}_n||_1.
\end{align}%
Also, we use identity loss to preserve source image's identity:
\begin{align}
   \mathcal{L}_{id} =& (1-cos(\upsilon_{id}(\boldsymbol{I}_s),\upsilon_{id}(\boldsymbol{I}_r))) +\\
   &(1-cos(\upsilon_{id}(\boldsymbol{I}_s),\upsilon_{id}(\boldsymbol{I}_{low}))),\nonumber
\end{align}%
where $\upsilon_{id}$ means the identity vectors generated by $\boldsymbol{F}_{id}$, and $cos(,)$ means the cosine similarity of two vectors. 
Finally, our SID loss is formulated as:
\begin{align}
\mathcal{L_{\text {sid}}}=\lambda_{\text {shape}} \mathcal{L}_{\text {shape}}+\lambda_{\text {id}} \mathcal{L}_{\text {id}},
\end{align}%
where $\lambda_{\text {id}}$ = $5$ and $\lambda_{\text {shape}}$ = $0.5$.

\paragraph{Realism Loss.} 
Realism loss contains segmentation loss, reconstruction loss, cycle loss, perceptual loss, and adversarial loss. 
Specifically, $\boldsymbol{M}_{low}$ and $\boldsymbol{M}_{r}$ in the SFF module are both under the guidance of a SOTA face segmentation network HRNet~\cite{sun2019high}. We dilated the masks of the target image to eliminate the limitation in face shape change and get $\boldsymbol{M}_{tar}$. 
The segmentation loss is formulated as:
\begin{align}
   \mathcal{L}_{seg} =   ||R(\boldsymbol{M}_{tar}) - \boldsymbol{M}_{low}||_1 + ||\boldsymbol{M}_{tar} - \boldsymbol{M}_{r}||_1,
\end{align}%
where $R(.)$ means the resize operation.

If $\boldsymbol{I}_s$ and $\boldsymbol{I}_t$ share the same identity, the predicted image should be the same as $\boldsymbol{I}_t$. So we use reconstruction loss to give pixel-wise supervision:
\begin{align}
   \mathcal{L}_{rec} =  ||\boldsymbol{I}_{r} - \boldsymbol{I}_{t}||_1 + ||\boldsymbol{I}_{low} - R(\boldsymbol{I}_{t})||_1.
\end{align}%

The cycle process can be conducted in the face swapping task too. Let ${\boldsymbol{I}_r}$ as the re-target image and the original target image as the re-source image. In the cycle process, we hope to generate results with re-source image's identity and re-target image's attributes, which means it should be the same as the original target image. The cycle loss is a supplement of pixel supervision and can help generate high-fidelity results:
\begin{align}
   \mathcal{L}_{cyc} =  ||{\boldsymbol{I}_t} - G({\boldsymbol{I}_r}, {\boldsymbol{I}_t})||_1,
\end{align}%
where $G$ means the whole generator of HifiFace.

To capture fine details and further improve the realism, we follow the Learned Perceptual Image Patch Similarity (LPIPS) loss in~\cite{zhang2018unreasonable} and adversarial objective in~\cite{choi2020stargan}.
Thus, our realism loss is formulated as:
\begin{align}
\mathcal{L_{\text {real}}}=\mathcal{L}_{\text {adv}}+\lambda_{\text {0}} \mathcal{L}_{\text {seg}}+\lambda_{\text {1}} \mathcal{L}_{\text {rec}}+\lambda_{\text {2}} \mathcal{L}_{\text {cyc}}+\lambda_{\text {3}} \mathcal{L}_{\text {lpips}},
\end{align}%
where $\lambda_{\text {0}}$ = $100$, $\lambda_{\text {1}}$ = $20$, $\lambda_{\text {2}}$ = $1$ and $\lambda_{\text {3}}$ = $5$.

\begin{figure}[t] 
\begin{center} 
\includegraphics[width=0.95\linewidth]{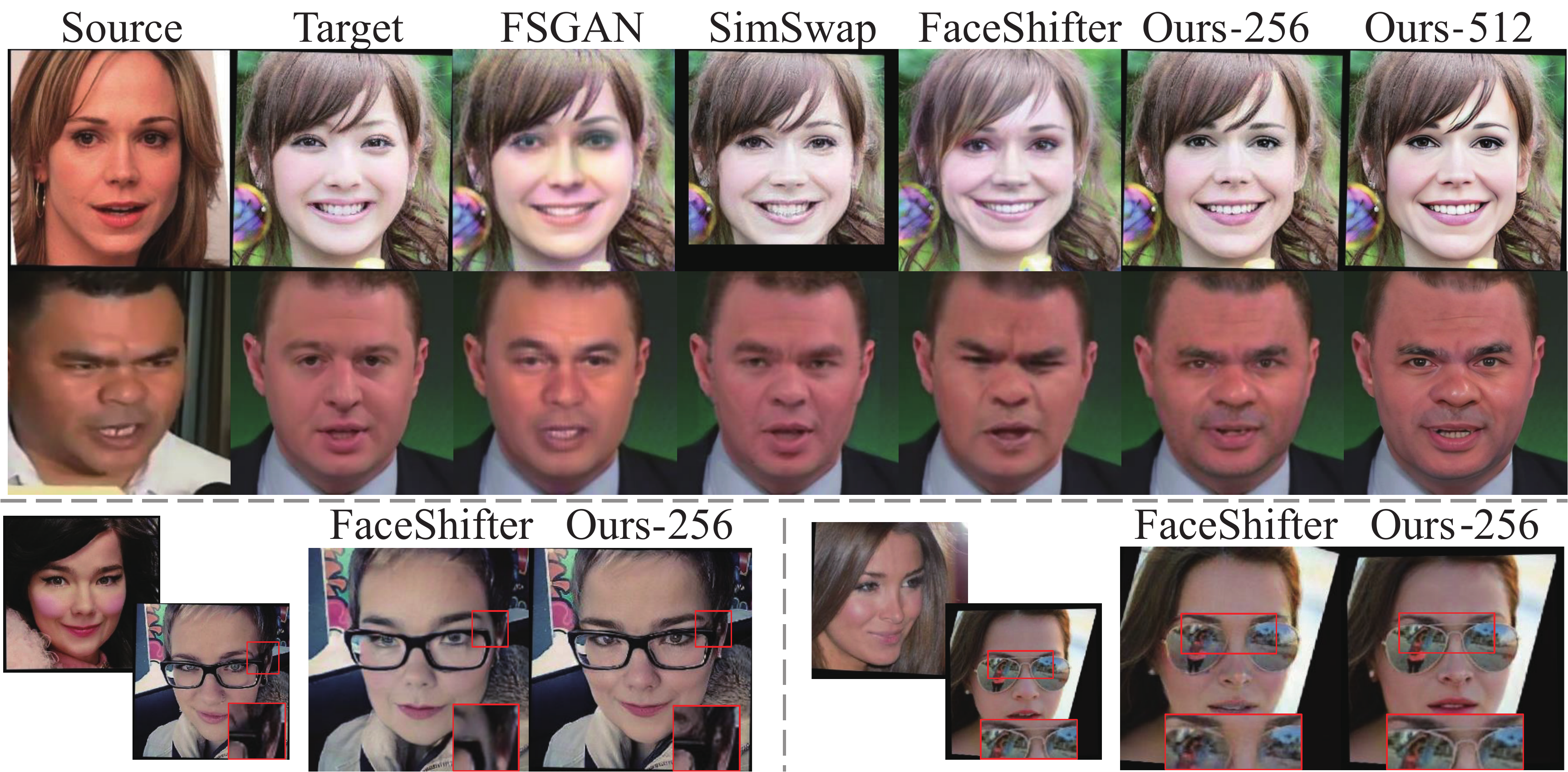} 
\end{center} 
\vspace{-2mm}
\caption{Comparison with FSGAN, SimSwap and FaceShifter. Our results can well preserve the source face shape, target attributes and have higher image quality, even when handling occlusion cases.} 
\label{fig:compare_big} 
\end{figure}

\paragraph{Overall Loss.}Our full loss is summarized as follows:
\begin{align}
\mathcal{L}=
\mathcal{L}_{\text {sid}}+
 \mathcal{L}_{\text {real}}.
\end{align}%

\section{Experiments}

\begin{table}[t]
\centering
\small
\begin{tabular}{cccc|cc}
\toprule
Method & ID$\uparrow$ & Pose$\downarrow$ & Shape$\downarrow$ &MAC~$\downarrow$ & FPS$\uparrow$      \\
\midrule
FaceSwap    & $54.19$  & $2.51$ & $0.610$ & - & -      \\
FaceShifter     & $97.38$  & $2.96$ & $0.511$ & $121.79$ & $22.34$     \\
SimSwap   & $92.83$  & $\textbf{1.53}$ & $0.540$ & $\textbf{55.69}$ & $\textbf{31.17}$    \\
\midrule 
Ours-$256$     & $\textbf{98.48}$  & $2.63$ & $\textbf{0.437}$ & $102.39$ & $25.29$   \\
\bottomrule
\end{tabular}
\vspace{-2mm}
\caption{Quantitative Experiments on FaceForensics++. FPS is tested under GPU V$100$.}
\label{tab:qua}
\end{table}

\begin{figure}[t] 
\begin{center} 
\includegraphics[width=0.92\linewidth]{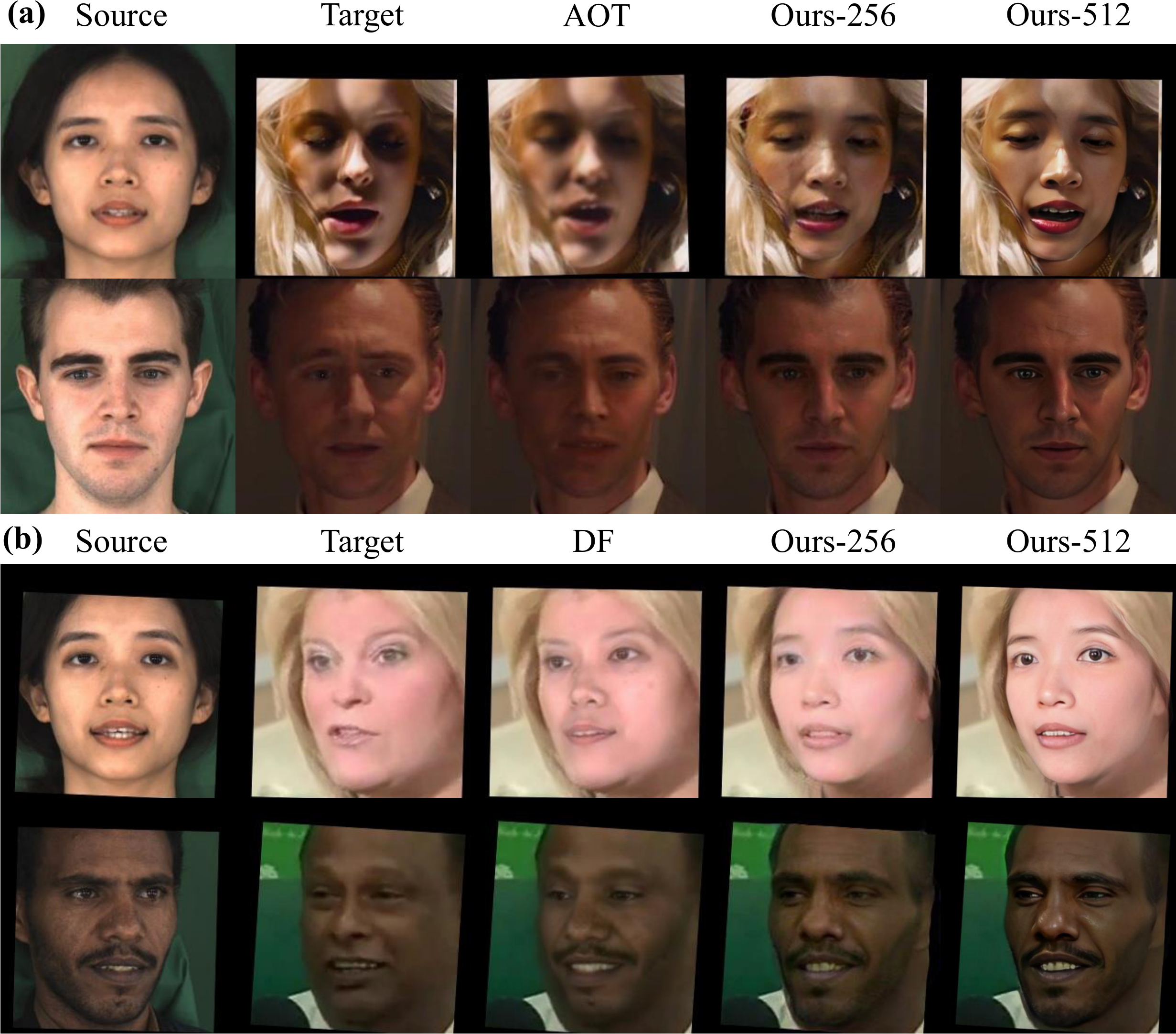} 
\end{center} 
\vspace{-2mm}
\caption{(a) Comparison with AOT. (b) Comparison with DF.} 
\label{fig:compare_aot_df} 
\end{figure}

\paragraph{Implementation Details.} 
We choose VGGFace$2$~\cite{cao2018vggface2} and Asian-Celeb~\cite{DeepGlint} as the training set. 
For our model with resolution $256$ (\textit{i.e.}, Ours-$256$), we remove images with either size smaller than $256$ for better image quality. 
For each image, we align the face using $5$ landmarks and crop to $256$$\times$$256$~\cite{li2019faceshifter}, which contains the whole face and some background regions. 
For our more precise model (\textit{i.e.}, Ours-$512$), we adopt a portrait enhancement network~\cite{li2020blind} to improve the resolution of the training images to $512$$\times$$512$ as supervision, and also correspondingly add another res-block in $\boldsymbol{F}_{up}$ of SFF compared to Ours-$256$. 
The ratio of training pairs with the same identity is $50$\%. ADAM~\cite{kingma2014adam} is used with $\beta1$ = $0$; $\beta2$ = $0.99$ and learning rate = $0.0001$. 
The model is trained with $200$K steps, using $4$ V$100$ GPUs and $32$ batch size.

\subsection{Qualitative Comparisons}
First, we compare our method with FSGAN~\cite{Nirkin2019fsgan}, SimSwap~\cite{chen2020simswap} and FaceShifter~\cite{li2019faceshifter} in Figure~\ref{fig:compare_big},  AOT~\cite{zhu2020aot} and DeeperForensics (DF)~\cite{jiang2020deeperforensics} in Figure~\ref{fig:compare_aot_df}.

As shown in Figure~\ref{fig:compare_big}, FSGAN shares the same face shape with target faces and it can not well transfer the lighting of the target image either.
SimSwap can not well preserve the identity of the source image especially for the face shape because it uses a feature matching loss and focuses more on the attributes. 
FaceShifter exhibits a strong identity preservation ability,
but it has two limitations: (1) Attribute recovery, while our HifiFace can well preserve all the attributes like face color, expression, and occlusion. (2) Complex framework with two stages, while HifiFace presents a more elegant end-to-end framework with even better recovered images. 
As shown in Figure~\ref{fig:compare_aot_df}(a), AOT is specially designed to overcome the lighting problem but is weak in identity similarity and fidelity.  As shown in Figure~\ref{fig:compare_aot_df}(b), DF has reduced the bad cases of style mismatch, but is weak in identity similarity too.
In contrast, our HifiFace not only perfectly preserves the lighting and face style, but also well captures the face shape of the source image and generates high quality swapped faces.
More results can be found in our $\mathsf{supplementary~material}$.

\begin{table}[t]
\centering
\small
\begin{tabular}{ccccc}
\toprule
\multirow{2}{*}[-0.2em]{Method} & \multicolumn{2}{c}{FF++} & \multicolumn{2}{c}{DFDC}\\
\cmidrule(l{2pt}r{2pt}){2-3}
\cmidrule(l{2pt}r{2pt}){4-5}
& AUC↓ & AP↓ & AUC↓ & AP↓   \\
\midrule
FaceSwap      & $66.83$  & $65.99$ & $92.30$ & $92.54$      \\
FaceShifter   & $41.62$  & $42.92$ & $77.18$ & $76.50$     \\
SimSwap   & $76.44$  & $72.63$ & $78.80$ & $78.44$    \\
\midrule
Ours-$256$     & $\textbf{38.97}$  & $\textbf{41.54}$ & $\textbf{62.29}$ & $\textbf{59.99}$     \\
\bottomrule
\end{tabular}
\vspace{-2mm}
\caption{Results in terms of AUC and AP on FF++ and DFDC.}
\label{tab:auc}
\end{table}

\begin{figure}[t] 
\begin{center} 
\includegraphics[width=0.9\linewidth]{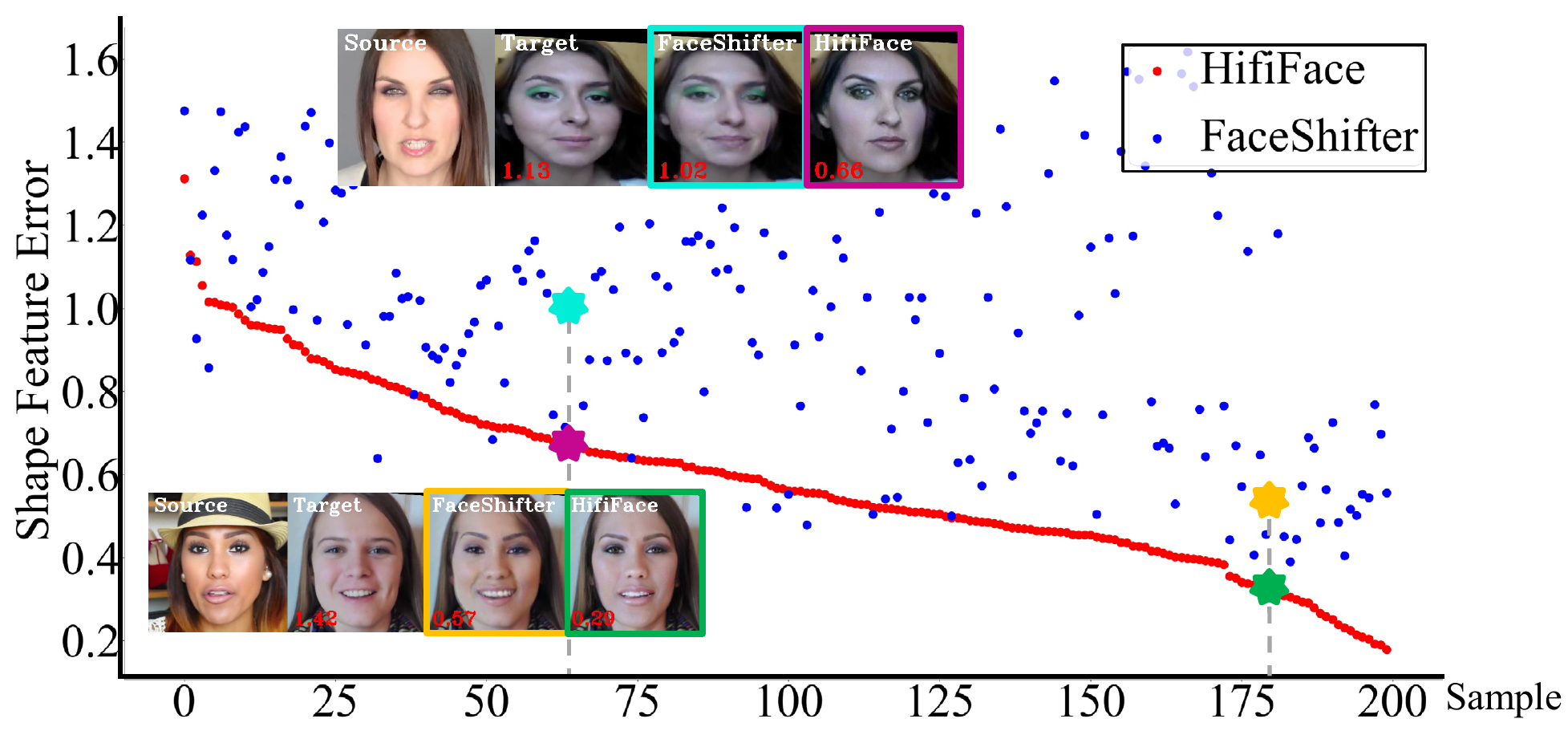} 
\end{center} 
\vspace{-2mm}
\caption{Face shape error between $I_r$ and $I_s$ in $200$ FF++ pairs with large shape difference. Samples are sorted by shape error of HifiFace. Same column index indicates the same source/target pair.}
\label{compare_shape} 
\end{figure}

\subsection{Quantitative Comparisons}

Next, we conduct quantitative comparison on FaceForensics (FF)++ ~\cite{rossler2019faceforensics++} dataset with respect to the following metrics: ID retrieval, pose error, face shape error, and performance on face forgery detection algorithms to again demonstrate the effectiveness of our HifiFace. 
For FaceSwap~\cite{FaceSwap} and FaceShifter, we evenly sample $10$ frames from each video and compose a $10$K test set. For SimSwap and our HifiFace, we generate face swapping results with the same source and target pairs above. 

For ID retrieval and pose error, we follow the same setting in ~\cite{li2019faceshifter,chen2020simswap}. 
As shown in Table~\ref{tab:qua}, HifiFace achieves the best ID retrieval score and is comparable with others in pose preservation.
For face shape error, we use another $3$D face reconstruction model~\cite{sanyal2019learning} to regress coefficients of each test face. The error is computed by L$2$ distances of identity coefficients between the swapped face and its source face, and our HifiFace achieves the lowest face shape error. 
The parameter and speed comparisons are also shown in Table~\ref{tab:qua}, and our HifiFace is faster than FaceShifter, along with higher generation quality. 

To further illustrate the ability of HifiFace in controlling the face shape, we visualize the samplewise shape differences between HifiFace and FaceShifter~\cite{li2019faceshifter} in Figure~\ref{compare_shape}. 
The results show that, when the source and target differs much in face shape, HifiFace significantly outperforms Faceshifter with $\textbf{95}$\% samples having smaller shape errors.


Besides, we apply the models from
FF++~\cite{rossler2019faceforensics++} and DeepFake Detection Challenge (DFDC)~\cite{dolhansky2019deepfake,dfdc} to examine the realism performance of HifiFace. 
The test set contains $10$K swapped faces and $10$K real faces from FF++ for each method. 
As shown in Table~\ref{tab:auc}, HifiFace achieves the best score, indicating higher fidelity to further help improve face forgery detection.

\begin{figure}[t] 
\begin{center} 
\includegraphics[width=0.9\linewidth]{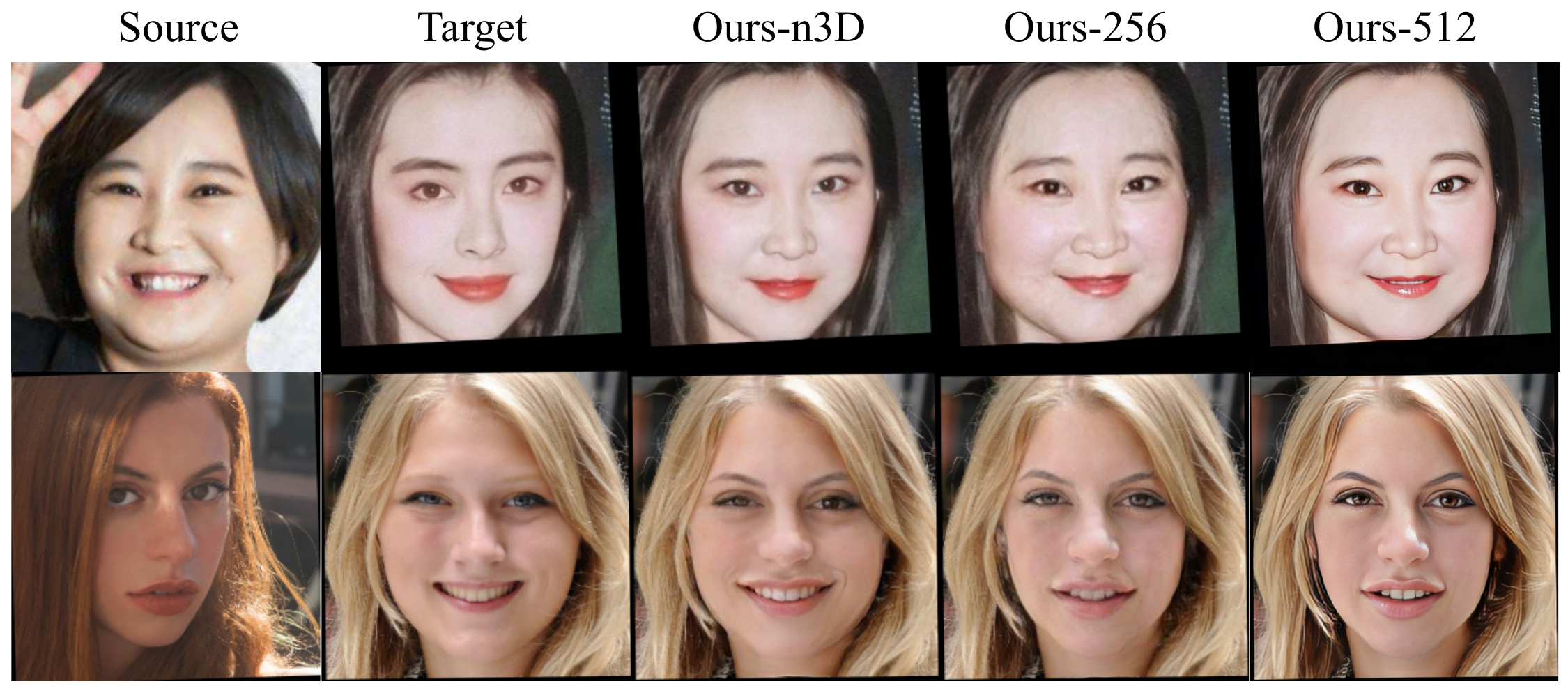} 
\end{center} 
\vspace{-2mm}
\caption{Ablation study for 3D shape-aware identity extractor.} 
\label{ab_3D}
\end{figure}

\begin{figure}[t] 
\begin{center} 
\includegraphics[width=0.9\linewidth]{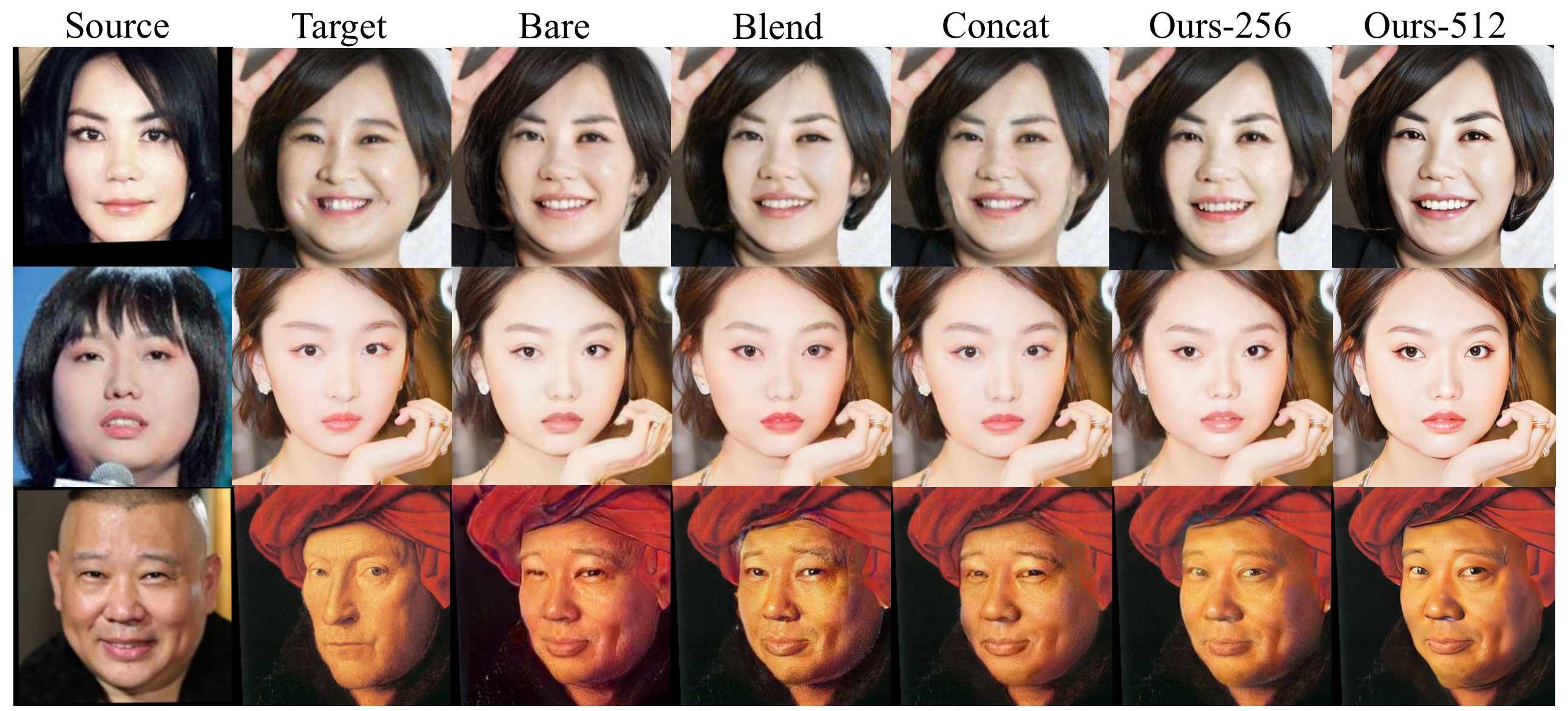} 
\end{center} 
\vspace{-2mm}
\caption{Ablation study for SFF module.} 
\label{ab_SFF} 
\end{figure}

\begin{figure}[t] 
\begin{center} 
\includegraphics[width=0.96\linewidth]{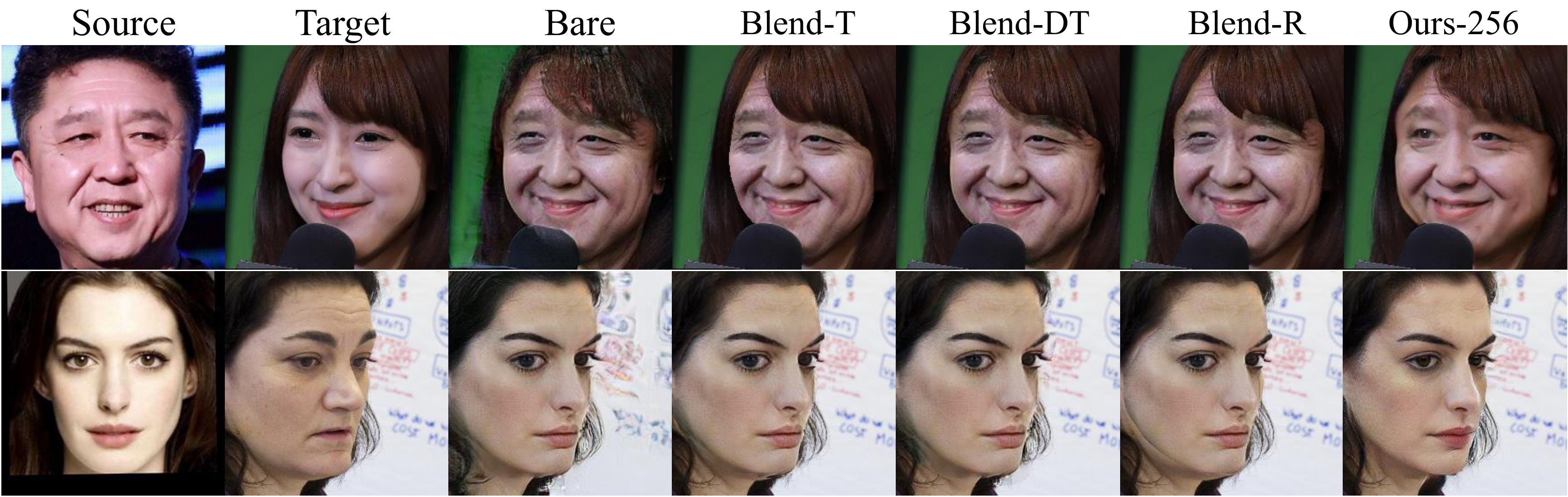} 
\end{center} 
\vspace{-2mm}
\caption{Comparison with results using directly mask blending. `Blend-T', `Blend-DT', and `Blend-R' mean blending bare results to the target image by the mask of target, the dilated mask of target and the mask of bare results, respectively. 
} 
\label{ab_face} 
\end{figure}

\begin{figure}[t] 
\begin{center} 
\includegraphics[width=0.96\linewidth]{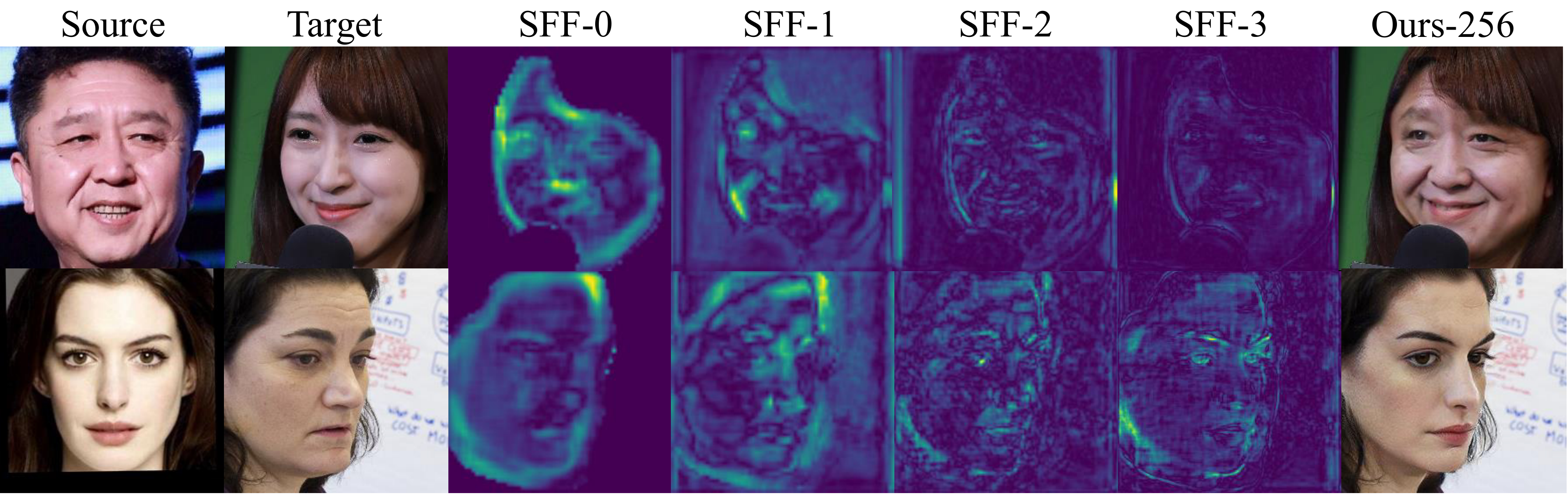} 
\end{center} 
\vspace{-2mm}
\caption{Difference feature maps of SFF.}
\label{ab_heatmap} 
\end{figure}

\subsection{Analysis of HifiFace}

\paragraph{3D Shape-Aware Identity.}
To verify the effectiveness of shape supervision $\mathcal{L}_{shape}$ on face shape, we train another model Ours-n$3$d, which replaces the shape-aware identity vector with the normal identity vector from $\boldsymbol{F}_{id}$. 
As shown in Figure~\ref{ab_3D}, the results of Ours-n$3$d can hardly change the face shape or have obvious artifacts, while the results of Ours-$256$ can generate results with much more similar face shape.

\paragraph{Semantic Facial Fusion.}
To verify the necessity of the SFF module, we compare with $3$ baseline models: 
($1$) `Bare' that removes both the feature-level and image-level fusion. 
($2$) `Blend' that removes the feature-level fusion. 
($3$) `Concat' that replaces the feature-level fusion with a concatenate. As shown in Figure~\ref{ab_SFF}, `Bare' can not well preserve the background and occlusion, `Blend' lacks legibility, and 'Concat' is weak in identity similarity, which proves that the SFF module can help preserve the attribute and improve the image quality without harming the identity.

\paragraph{Face Shape Preservation in Face Swapping.} 
Face shape preservation is quite difficult for face swapping, which is not just because of the difficulty in getting shape information, but also the challenge of inpainting when the face shape has changed. 
Blending is a valid way to preserve occlusion and background, but it is hard to be applied when the face shape changes. 
As shown in Figure~\ref{ab_face}, when the source face is fatter than the target face (row $1$), it may limit the change of face shape in Blend-T. If we use Blend-DT or Blend-R, it can not well handle the occlusion. 
When the source face is thinner than the target (row $2$), it is easy to bring artifacts around the face in Blend-T and Blend-DT and may cause a double face in Blend-R. 
In contrast, our HifiFace can apply the blending without the above issue, because our SFF module has the ability to inpaint the edge of the predicted mask. 

To further illustrate how SFF addresses the problem, we show difference feature maps of every stage in the SFF module, named SFF-$0$$\sim$$3$, between the input of ($I_s$,$I_t$) and ($I_t$,$I_t$), where ($I_s$,$I_t$) obtains Ours-$256$ and ($I_t$,$I_t$) achieves target itself. 
In Figure~\ref{ab_heatmap}, the bright area means where the face shape changes or contains artifacts. 
SFF module recombines the feature between the face region and non-face area and focuses more on the contour of the predicted mask, which brings great benefits for inpainting areas where the shape changes.

\section{Conclusions}
In this work, we propose a high fidelity face swapping method, named HifiFace, which can well preserve the face shape of the source face and generate photo-realistic results. A 3D shape-aware identity extractor is proposed to help preserve identity including face shape. An SFF module is proposed to achieve a better combination in feature-level and image-level for realistic image generation. Extensive experiments demonstrate that our method can generate higher fidelity results than previous SOTA face swapping methods both quantitatively and qualitatively. Last but not least, HifiFace can also be served as a sharp spear, which contributes to the development of the face forgery detection community.


\small 
\bibliographystyle{named}
\bibliography{ijcai21}

\begin{thebibliography}{}

\bibitem[\protect\citeauthoryear{Alexander \bgroup \em et al.\egroup
  }{2009}]{alexander2009creating}
Oleg Alexander, Mike Rogers, William Lambeth, Matt Chiang, and Paul Debevec.
\newblock Creating a photoreal digital actor: The digital emily project.
\newblock In {\em 2009 Conference for Visual Media Production}, pages 176--187.
  IEEE, 2009.

\bibitem[\protect\citeauthoryear{Blanz and Vetter}{1999}]{blanz1999morphable}
Volker Blanz and Thomas Vetter.
\newblock A morphable model for the synthesis of 3d faces.
\newblock In {\em Proceedings of the 26th annual conference on Computer
  graphics and interactive techniques}, pages 187--194, 1999.

\bibitem[\protect\citeauthoryear{Cao \bgroup \em et al.\egroup
  }{2018}]{cao2018vggface2}
Qiong Cao, Li~Shen, Weidi Xie, Omkar~M Parkhi, and Andrew Zisserman.
\newblock Vggface2: A dataset for recognising faces across pose and age.
\newblock In {\em FG}, pages 67--74. IEEE, 2018.

\bibitem[\protect\citeauthoryear{Chen \bgroup \em et al.\egroup
  }{2020}]{chen2020simswap}
Renwang Chen, Xuanhong Chen, Bingbing Ni, and Yanhao Ge.
\newblock Simswap: An efficient framework for high fidelity face swapping.
\newblock In {\em Proceedings of the 28th ACM International Conference on
  Multimedia}, pages 2003--2011, 2020.

\bibitem[\protect\citeauthoryear{Choi \bgroup \em et al.\egroup
  }{2020}]{choi2020stargan}
Yunjey Choi, Youngjung Uh, Jaejun Yoo, and Jung-Woo Ha.
\newblock Stargan v2: Diverse image synthesis for multiple domains.
\newblock In {\em Proceedings of the IEEE Conference on Computer Vision and
  Pattern Recognition}, pages 8188--8197, 2020.

\bibitem[\protect\citeauthoryear{DeepGlint}{2020}]{DeepGlint}
DeepGlint.
\newblock http://trillionpairs.deepglint.com.
\newblock {\em Accessed: 2020-12-20}, 2020.

\bibitem[\protect\citeauthoryear{Deng \bgroup \em et al.\egroup
  }{2019}]{deng2019accurate}
Yu~Deng, Jiaolong Yang, Sicheng Xu, Dong Chen, Yunde Jia, and Xin Tong.
\newblock Accurate 3d face reconstruction with weakly-supervised learning: From
  single image to image set.
\newblock In {\em Proceedings of the IEEE Conference on Computer Vision and
  Pattern Recognition Workshops}, pages 0--0, 2019.

\bibitem[\protect\citeauthoryear{Dolhansky \bgroup \em et al.\egroup
  }{2019}]{dolhansky2019deepfake}
Brian Dolhansky, Russ Howes, Ben Pflaum, Nicole Baram, and Cristian~Canton
  Ferrer.
\newblock The deepfake detection challenge (dfdc) preview dataset.
\newblock {\em arXiv preprint arXiv:1910.08854}, 2019.

\bibitem[\protect\citeauthoryear{FaceSwap}{}]{FaceSwap}
FaceSwap.
\newblock https://github.com/ondyari/faceforensics/
  tree/master/dataset/faceswapkowalski.
\newblock {\em Accessed: 2020-12-20}.

\bibitem[\protect\citeauthoryear{Goodfellow \bgroup \em et al.\egroup
  }{2014}]{goodfellow2014generative}
Ian Goodfellow, Jean Pouget-Abadie, Mehdi Mirza, Bing Xu, David Warde-Farley,
  Sherjil Ozair, Aaron Courville, and Yoshua Bengio.
\newblock Generative adversarial nets.
\newblock In {\em Advances in neural information processing systems}, pages
  2672--2680, 2014.

\bibitem[\protect\citeauthoryear{He \bgroup \em et al.\egroup
  }{2016}]{he2016deep}
Kaiming He, Xiangyu Zhang, Shaoqing Ren, and Jian Sun.
\newblock Deep residual learning for image recognition.
\newblock In {\em Proceedings of the IEEE Conference on Computer Vision and
  Pattern Recognition}, pages 770--778, 2016.

\bibitem[\protect\citeauthoryear{Huang \bgroup \em et al.\egroup
  }{2020}]{huang2020curricularface}
Yuge Huang, Yuhan Wang, Ying Tai, Xiaoming Liu, Pengcheng Shen, Shaoxin Li,
  Jilin Li, and Feiyue Huang.
\newblock Curricularface: adaptive curriculum learning loss for deep face
  recognition.
\newblock In {\em Proceedings of the IEEE Conference on Computer Vision and
  Pattern Recognition}, pages 5901--5910, 2020.

\bibitem[\protect\citeauthoryear{Isola \bgroup \em et al.\egroup
  }{2017}]{isola2017image}
Phillip Isola, Jun-Yan Zhu, Tinghui Zhou, and Alexei~A Efros.
\newblock Image-to-image translation with conditional adversarial networks.
\newblock In {\em Proceedings of the IEEE Conference on Computer Vision and
  Pattern Recognition}, pages 1125--1134, 2017.

\bibitem[\protect\citeauthoryear{Jiang \bgroup \em et al.\egroup
  }{2020}]{jiang2020deeperforensics}
Liming Jiang, Ren Li, Wayne Wu, Chen Qian, and Chen~Change Loy.
\newblock Deeperforensics-1.0: A large-scale dataset for real-world face
  forgery detection.
\newblock In {\em Proceedings of the IEEE Conference on Computer Vision and
  Pattern Recognition}, pages 2886--2895. IEEE, 2020.

\bibitem[\protect\citeauthoryear{Karras \bgroup \em et al.\egroup
  }{2019}]{karras2019style}
Tero Karras, Samuli Laine, and Timo Aila.
\newblock A style-based generator architecture for generative adversarial
  networks.
\newblock In {\em Proceedings of the IEEE Conference on Computer Vision and
  Pattern Recognition}, pages 4401--4410, 2019.

\bibitem[\protect\citeauthoryear{Kingma and Ba}{2014}]{kingma2014adam}
Diederik~P Kingma and Jimmy Ba.
\newblock Adam: A method for stochastic optimization.
\newblock {\em arXiv preprint arXiv:1412.6980}, 2014.

\bibitem[\protect\citeauthoryear{Li \bgroup \em et al.\egroup
  }{2019}]{li2019faceshifter}
Lingzhi Li, Jianmin Bao, Hao Yang, Dong Chen, and Fang Wen.
\newblock Faceshifter: Towards high fidelity and occlusion aware face swapping.
\newblock {\em arXiv preprint arXiv:1912.13457}, 2019.

\bibitem[\protect\citeauthoryear{Li \bgroup \em et al.\egroup
  }{2020}]{li2020blind}
Xiaoming Li, Chaofeng Chen, Shangchen Zhou, Xianhui Lin, Wangmeng Zuo, and Lei
  Zhang.
\newblock Blind face restoration via deep multi-scale component dictionaries.
\newblock In {\em European Conference on Computer Vision}, pages 399--415.
  Springer, 2020.

\bibitem[\protect\citeauthoryear{Liu \bgroup \em et al.\egroup
  }{2019}]{liu2019identity}
Jialun Liu, Wenhui Li, Hongbin Pei, Ying Wang, Feng Qu, You Qu, and Yuhao Chen.
\newblock Identity preserving generative adversarial network for cross-domain
  person re-identification.
\newblock {\em IEEE Access}, 7:114021--114032, 2019.

\bibitem[\protect\citeauthoryear{Natsume \bgroup \em et al.\egroup
  }{2018}]{natsume2018fsnet}
Ryota Natsume, Tatsuya Yatagawa, and Shigeo Morishima.
\newblock Fsnet: An identity-aware generative model for image-based face
  swapping.
\newblock In {\em Asian Conference on Computer Vision}, pages 117--132.
  Springer, 2018.

\bibitem[\protect\citeauthoryear{Nirkin \bgroup \em et al.\egroup
  }{2018}]{nirkin2018face}
Yuval Nirkin, Iacopo Masi, Anh~Tran Tuan, Tal Hassner, and Gerard Medioni.
\newblock On face segmentation, face swapping, and face perception.
\newblock In {\em FG}, pages 98--105. IEEE, 2018.

\bibitem[\protect\citeauthoryear{Nirkin \bgroup \em et al.\egroup
  }{2019}]{Nirkin2019fsgan}
Yuval Nirkin, Yosi Keller, and Tal Hassner.
\newblock Fsgan: Subject agnostic face swapping and reenactment.
\newblock In {\em ICCV}, 2019.

\bibitem[\protect\citeauthoryear{Rossler \bgroup \em et al.\egroup
  }{2019}]{rossler2019faceforensics++}
Andreas Rossler, Davide Cozzolino, Luisa Verdoliva, Christian Riess, Justus
  Thies, and Matthias Nie{\ss}ner.
\newblock Faceforensics++: Learning to detect manipulated facial images.
\newblock In {\em Proceedings of the IEEE International Conference on Computer
  Vision}, pages 1--11, 2019.

\bibitem[\protect\citeauthoryear{Sanyal \bgroup \em et al.\egroup
  }{2019}]{sanyal2019learning}
Soubhik Sanyal, Timo Bolkart, Haiwen Feng, and Michael~J Black.
\newblock Learning to regress 3d face shape and expression from an image
  without 3d supervision.
\newblock In {\em Proceedings of the IEEE Conference on Computer Vision and
  Pattern Recognition}, pages 7763--7772, 2019.

\bibitem[\protect\citeauthoryear{selimsef}{2020}]{dfdc}
selimsef.
\newblock https://github.com/selimsef/dfdc\_deep- fake\_challenge.
\newblock {\em Accessed: 2021-01-10}, 2020.

\bibitem[\protect\citeauthoryear{Sun \bgroup \em et al.\egroup
  }{2019}]{sun2019high}
Ke~Sun, Yang Zhao, Borui Jiang, Tianheng Cheng, Bin Xiao, Dong Liu, Yadong Mu,
  Xinggang Wang, Wenyu Liu, and Jingdong Wang.
\newblock High-resolution representations for labeling pixels and regions.
\newblock {\em arXiv preprint arXiv:1904.04514}, 2019.

\bibitem[\protect\citeauthoryear{Thies \bgroup \em et al.\egroup
  }{2016}]{thies2016face2face}
Justus Thies, Michael Zollhofer, Marc Stamminger, Christian Theobalt, and
  Matthias Nie{\ss}ner.
\newblock Face2face: Real-time face capture and reenactment of rgb videos.
\newblock In {\em Proceedings of the IEEE Conference on Computer Vision and
  Pattern Recognition}, pages 2387--2395, 2016.

\bibitem[\protect\citeauthoryear{Zhang \bgroup \em et al.\egroup
  }{2018}]{zhang2018unreasonable}
Richard Zhang, Phillip Isola, Alexei~A Efros, Eli Shechtman, and Oliver Wang.
\newblock The unreasonable effectiveness of deep features as a perceptual
  metric.
\newblock In {\em Proceedings of the IEEE Conference on Computer Vision and
  Pattern Recognition}, pages 586--595, 2018.

\bibitem[\protect\citeauthoryear{Zhu \bgroup \em et al.\egroup
  }{2020}]{zhu2020aot}
Hao Zhu, Chaoyou Fu, Qianyi Wu, Wayne Wu, Chen Qian, and Ran He.
\newblock Aot: Appearance optimal transport based identity swapping for forgery
  detection.
\newblock {\em Advances in Neural Information Processing Systems}, 33, 2020.

\end{thebibliography}

\normalsize

\begin{figure*}[t] 
\begin{center} 
\includegraphics[width=0.95\linewidth]{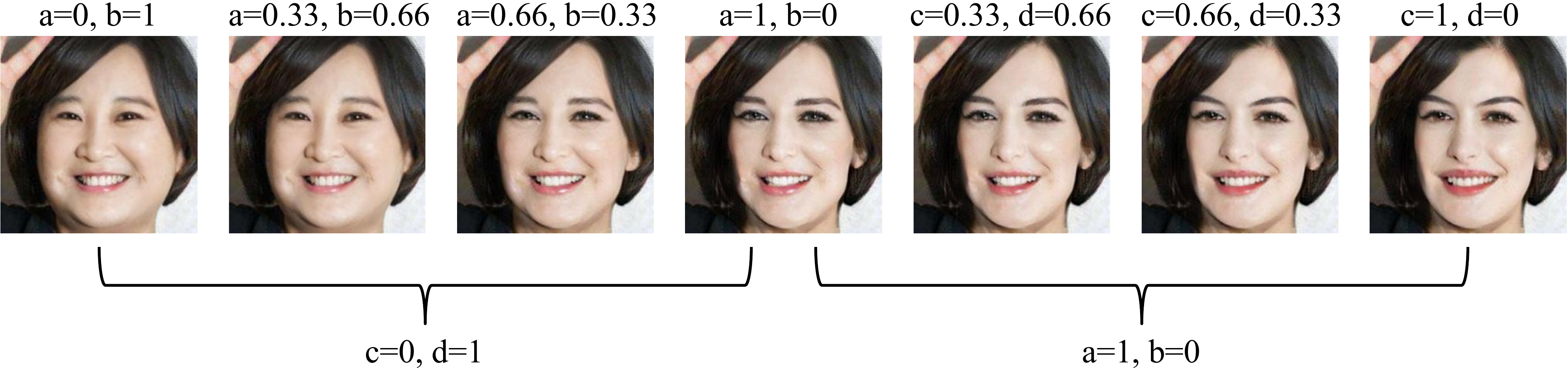} 
\end{center} 
\caption{Interpolated results with different compositions of SID.} 
\label{supp_sid} 
\end{figure*}

\section*{Network Structures} 
\begin{figure}[t] 
\begin{center} 
\includegraphics[width=0.96\linewidth]{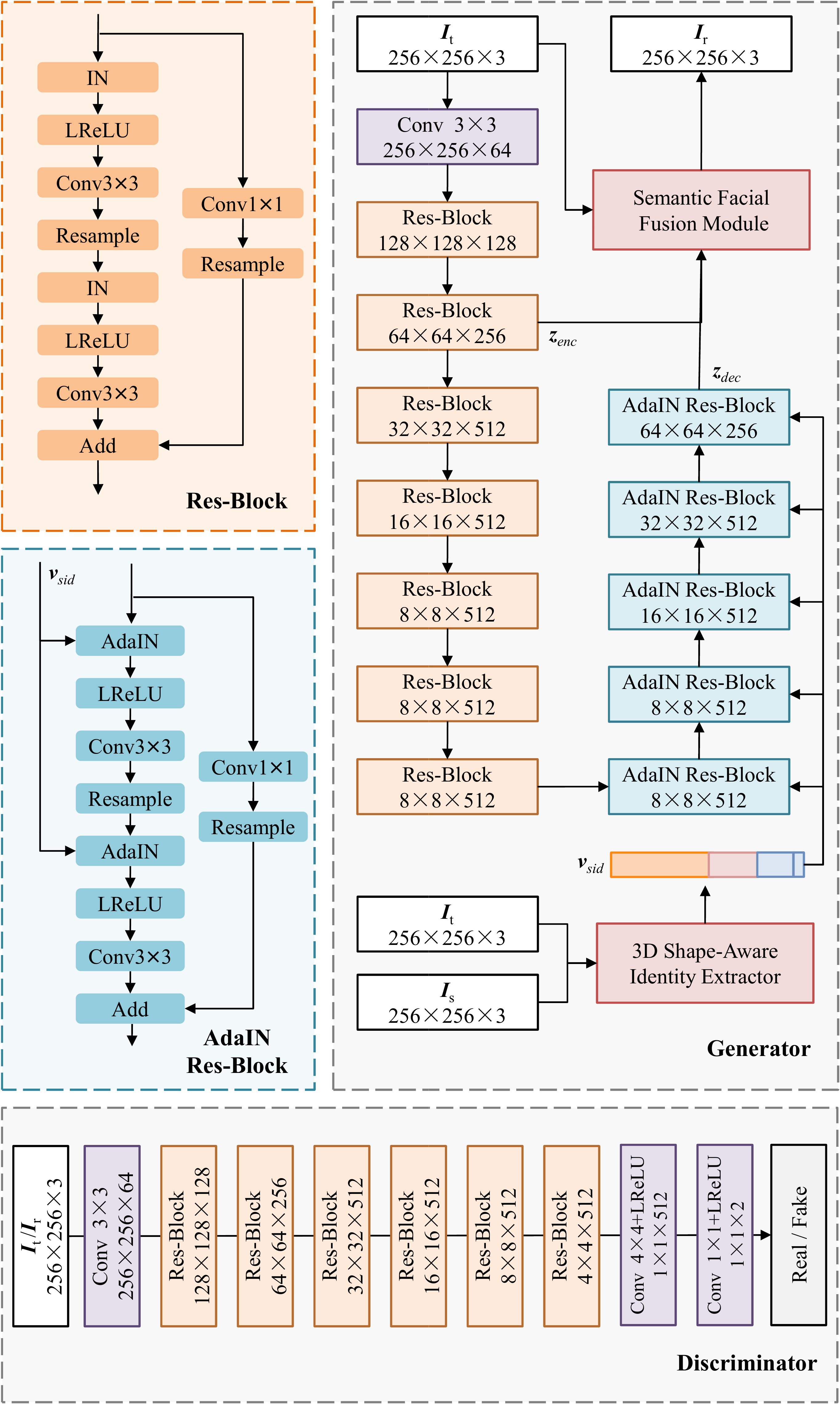} 
\end{center} 
\caption{Architectural details of HifiFace.}
\label{arch} 
\end{figure}

Detailed structures of our HifiFace are given in Figure~\ref{arch}. For all residual units, we use the Leaky ReLU (LReLU) as the activation function. Resample means the Average Pooling or the Upsampling, which is used to change the size of feature maps. Res-Blocks with the Instance
Normalization (IN) are used in encoder, while  Res-Blocks with the Adaptive Instance Normalization (AdaIN) are used in decoder.

\section*{More Results} 

To analyse the specific impacts of the shape information from 3D face reconstruction model and the identity information from face recognition model, we adjust the composition of SID to generate interpolated results. It is formulated as:
\begin{align}
   &\mathcal\psi_{in} = a \psi_{s} + b \psi_{t}, 
\end{align}%
\begin{align}
   &\mathcal\upsilon_{in} = c \upsilon_{s} + d \upsilon_{t}, 
\end{align}%
where $\psi_{s}$, $\psi_{t}$ and $\psi_{in}$ means the $3$D identity coefficients of source, target and interpolated image, $\upsilon_{s}$, $\upsilon_{t}$ and $\upsilon_{in}$ means source, target and interpolated image's identity vector from recognition model. 

As we can see in Figure~\ref{supp_sid} rows $1-4$, we first fix $c=0$ and $d=1$, the face shape can still change but lake of identity detail. Then in rows $4-7$, we fix $a=1$ and $b=0$, and the identity becomes more similar. The results prove that the shape information control the basic of shape and identity, while the identity vector is helpful to the identity texture.

\begin{figure*}[t] 
\begin{center} 
\includegraphics[width=0.95\linewidth]{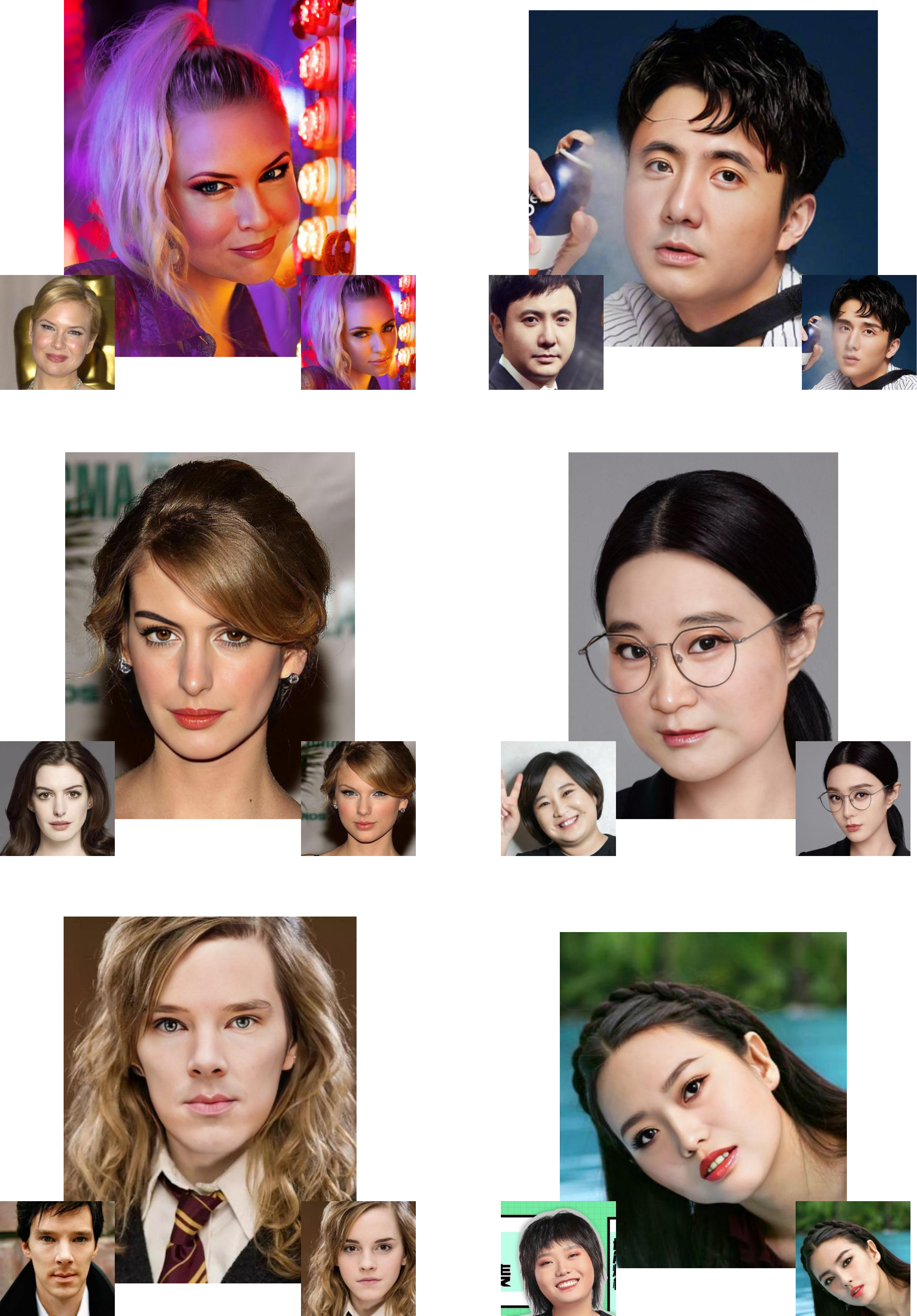} 
\end{center} 
\caption{More results on high resolution wild faces. 
The face in the target image is replaced by the face in the source image.}
\label{wild_1} 
\end{figure*}

\begin{figure*}[t] 
\hsize=\textwidth
\centering
\subfigure{
    \includegraphics[width=0.6\textwidth]{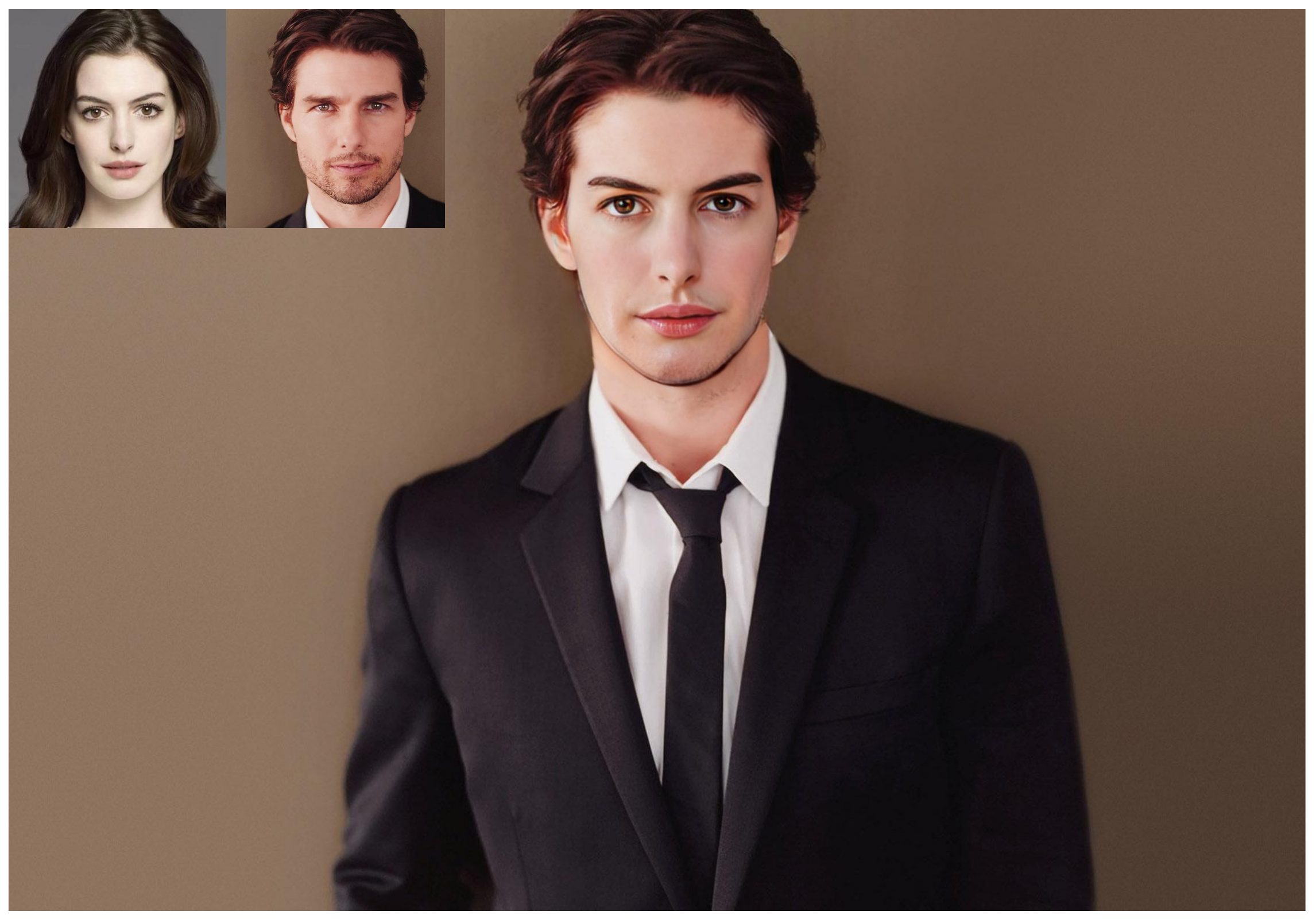}    
}

\subfigure{
    \includegraphics[width=0.6\textwidth]{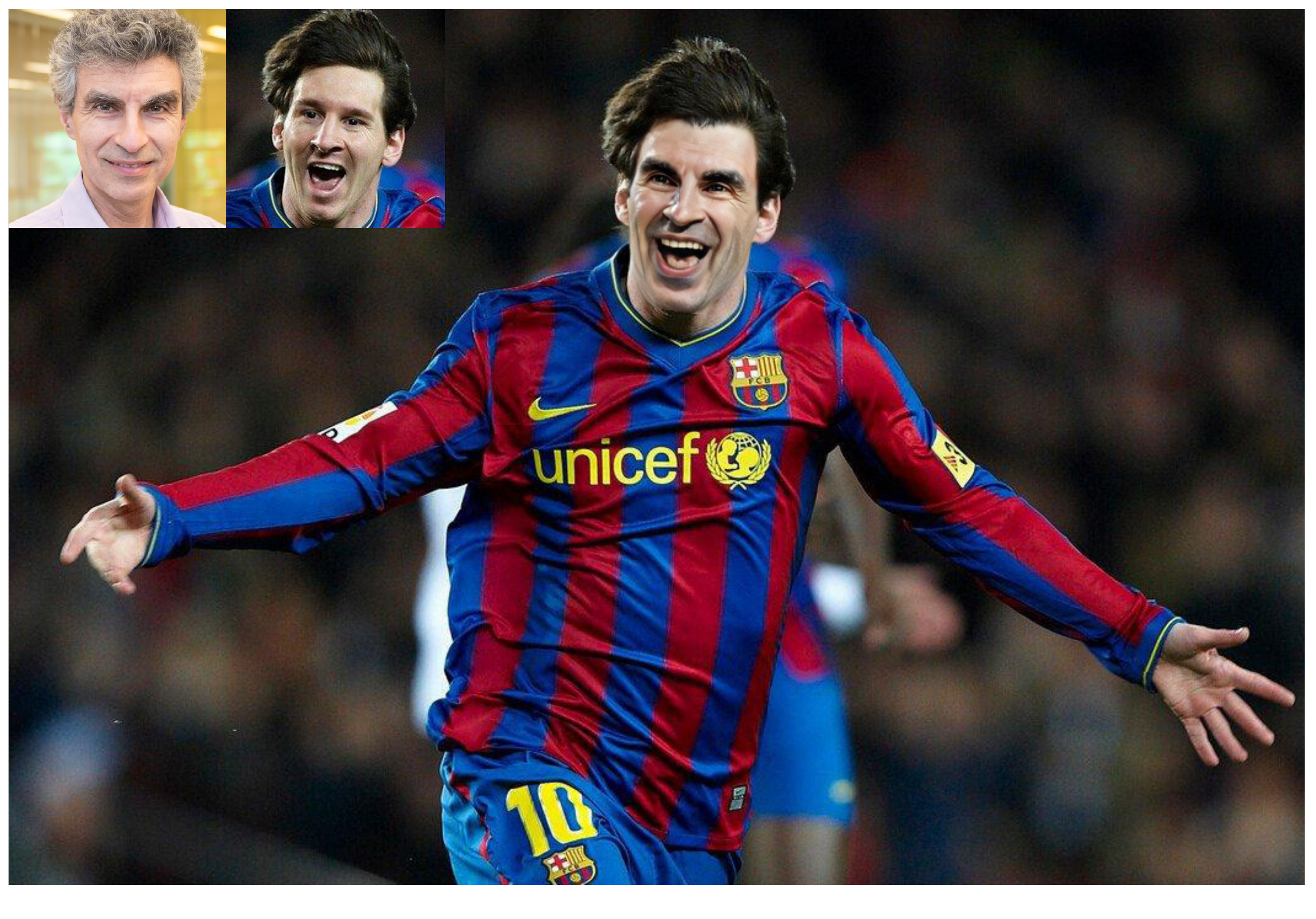}    
}

\subfigure{
    \includegraphics[width=0.6\textwidth]{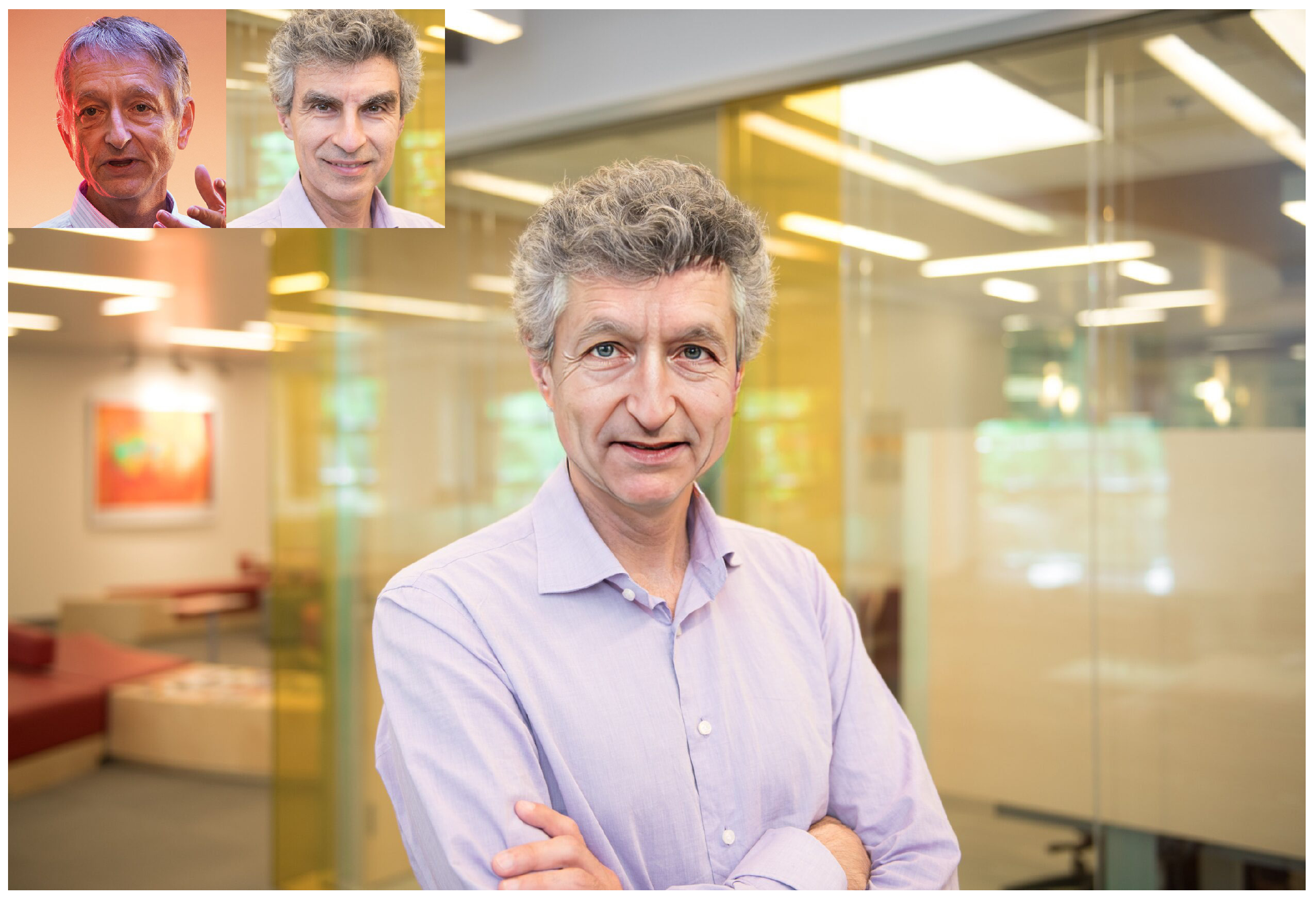}    
}

\caption{Some results on high quality real world photos. 
The face in the target image is replaced by the face in the source image. This is to show that the faces generated by our method can be very naturally integrated into high-resolution shot real world scenes.} 
\label{wild_2}
\end{figure*}

In the end, we download lots of wild face images from Internet
and generate more face swapping results in Figure~\ref{wild_1} and Figure~\ref{wild_2} to demonstrate the strong capability of our methods. And more results can be found at{~\color{magenta} ~\url{https://johann.wang/HifiFace}}.

\newpage

\newpage

\end{document}